\documentclass[letterpaper]{article} 
\usepackage{aaai25}  
\usepackage{times}  
\usepackage{helvet}  
\usepackage{courier}  
\usepackage[hyphens]{url}  
\usepackage{graphicx} 
\usepackage{natbib}  
\usepackage{caption} 
\usepackage{algorithm}
\usepackage{algorithmic}
\usepackage{svg}
\usepackage{paralist}
\usepackage{newfloat}
\usepackage{listings}
\DeclareCaptionStyle{ruled}{labelfont=normalfont,labelsep=colon,strut=off} 
\floatstyle{ruled}
\newfloat{listing}{tb}{lst}{}
\floatname{listing}{Listing}
\title{HyperCausalLP: Causal Link Prediction using Hyper-Relational Knowledge Graph}

\author {
    Utkarshani Jaimini\textsuperscript{\rm 1},
    Cory Henson\textsuperscript{\rm 2},
    Amit Sheth\textsuperscript{\rm 1}
}
\affiliations {
    \textsuperscript{\rm 1}Artificial Intelligence Institute, University of South Carolina, Columbia, SC, USA\\
    \textsuperscript{\rm 2}Bosch Center for Artificial Intelligence, Pittsburgh, PA, USA\\
    ujaimini@email.sc.edu, coryhenson@us.bosch.com, amit@sc.edu
}

\usepackage{bibentry}

\begin{document}

\maketitle

\begin{abstract}
Causal networks are often incomplete with missing causal links. This is due to various issues, such as missing observation data. Recent approaches to the issue of incomplete causal networks have used knowledge graph link prediction methods to find the missing links. 
In the causal link \textit{A causes B causes C}, the influence of A to C is influenced by B which is known as a mediator. Existing approaches using knowledge graph link prediction do not consider these mediated causal links. This paper presents HyperCausalLP, an approach designed to find missing causal links within a causal network with the help of mediator links. The problem of missing links is formulated as a hyper-relational knowledge graph completion. 
The approach uses a knowledge graph link prediction model trained on a hyper-relational knowledge graph with the mediators. The approach is evaluated on a causal benchmark dataset, CLEVRER-Humans. Results show that the inclusion of knowledge about mediators in causal link prediction using hyper-relational knowledge graph improves the performance on an average by 5.94\% mean reciprocal rank.
\end{abstract}

%

\section{Introduction}
\begin{figure}[h!]
    \centering
    \includegraphics[width=0.9\columnwidth]{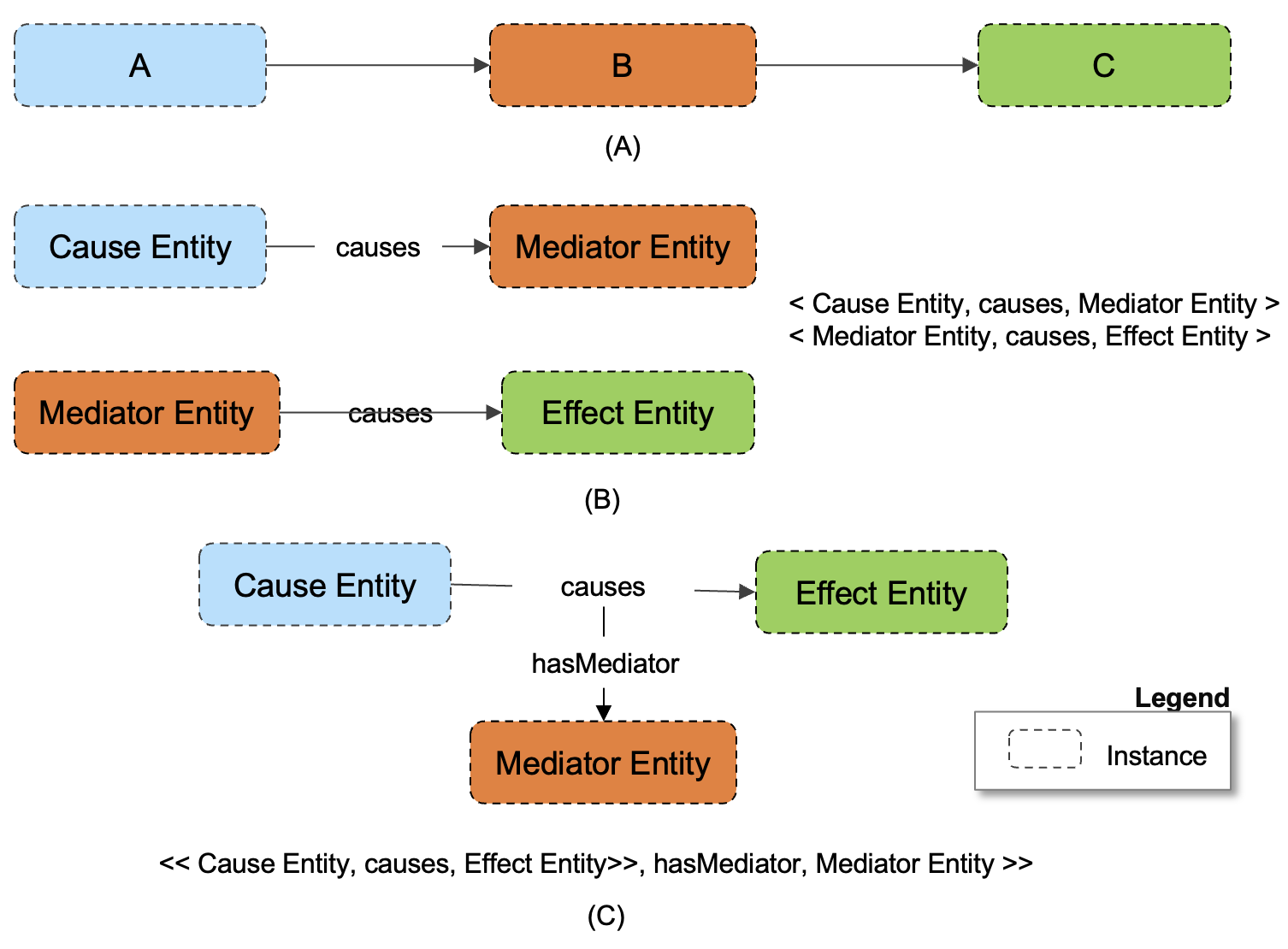}
    \caption{Causal link. (A) A serial causal connection where A causes B and eventually B causes C. The node A is known as a cause, C is known as an effect, and B is known as a mediator. (B) A serial causal link, the link is encoded as a knowledge graph link using RDF format, (C) Causal link as a hyper-relational link where the mediator entity is represented a hyper-relation with the hyper-relation predicate, hasMediator. The link is encoded as a knowledge graph link using RDF-Star format}
    \label{fig:causalLink}
\end{figure}

Causality is traditionally represented using a causal network, where the nodes represent events and edges represent the causal link between two events \cite{pearl2009causality}. Consider an example of a simple binary causal link: A causes B as shown in Figure \ref{fig:causalLink}(A). In this case, A is the cause, and B is the effect. Such causal links can also be chained together where A causes B and then B causes C. In a more complex case, there is a causal link between A and C that is mediated by B. The nodes A and C are called the cause and effect respectively, and the node B is called a mediator. A mediator helps in explaining the relationship between cause (independent node) and its effect (dependent node). It provides insights into the pathway linking cause and effect, capturing the contextual information. 
A complete network with all causal links is important for many downstream applications. In practice, however, causal networks are often incomplete with missing causal links. Recent approaches have successfully resolved this issue by encoding the causal network within a triple-based knowledge graph (i.e., Resource Description Framework (RDF) \cite{causalODP}) and then using knowledge graph link prediction techniques to find the missing causal links \cite{jaimini2024causaldisco}. 
While the existing approaches using knowledge graph (KG) link prediction can predict direct binary causal links, e.g., A causes B, they cannot predict the more complex mediated causal links, e.g., A causes C mediated by B. The mediated link captures the context information. 

In this paper, we present a Hyper Causal Link Prediction approach, HyperCausalLP\footnote{Code - https://github.com/CausalKG/HyperCausalLP/}), for finding the missing causal links in an incomplete causal network using hyper-relational KG link prediction. 
It uses hyper-relational causal knowledge graph (CausalKG) to represent the complex causal relations in the causal network. Figure \ref{fig:causalLink}(C) shows how mediated causal links is encoded as a hyper-relation. RDF-star\footnote{https://www.w3.org/2021/12/rdf-star.html} is used to encode these causal links \cite{causalODP}. 
The main contributions of this paper are:
\begin{enumerate}
\item A novel formulation of the task of finding missing causal links in an incomplete causal network as a hyper-relational KG completion problem.  
\item Incorporation of mediated links into causal link prediction, which leads to improved performance.
\item Demonstration of the approach for causal link prediction using a causal benchmark dataset.
\item Use of additional domain knowledge for evaluating causal link prediction.
\end{enumerate}


The hyper-relational CausalKG is transformed into a KG embedding (KGE) model using StarE \cite{StarEGalkin2020message} algorithm, which uses a neural network-based message-passing framework. 
This approach to finding missing causal links with mediators is evaluated using a causal benchmark dataset.
StarE based hyper-relational KG extend a triple representation with any number of qualifies. It separates the qualifier relation and entity from the main triple. It does not have an upper bound on the number of qualifier per triple.
The contributions of this paper are highlighted through the following four research questions: 
\begin{compactitem}
    \item \textbf{\textit{RQ1}}: Can the information contained in a causal network be effectively encoded into a hyper-relational causal KG?
    \item \textbf{\textit{RQ2}}: Can  KG completion techniques, i.e., link prediction, be harnessed to uncover missing causal links?
    \item \textbf{\textit{RQ3}}: Does the integration of mediated links lead to improvements in the performance of causal link prediction?
    \item \textbf{\textit{RQ4}}: Does the integration of additional domain knowledge lead to improvements in the performance of causal link prediction? 
\end{compactitem}

The rest of the paper proceeds as follows: Section 2 describes the related work. Section 3 defines the problem formulation, followed by Section 4, which details the methodology. Section 5 details the evaluation, with the results and discussion outlined in Section 6. Section 7 provides a conclusion with future direction.

\section{Related work}
\textit{Knowledge graph link prediction}  
The KG link prediction approach ranges from translation-based models, semantic matching models, and convolutional neural network-based models \cite{rossi2021knowledge, wang2017knowledgesurvey, wang2021survey}. These methods learn embedding for each entity and relation and use a scoring function to predict the likelihood of a triple being true.  The graph neural network-based methods use message-passing approaches utilizing semantically rich neighborhood information present in the KG \cite{vashishth2019composition, schlichtkrull2018modeling, nguyen2022node, mohamed2023locality,li2024evaluating}.  

\textit{Causal link prediction}
The existing techniques for causal link prediction typically focus on predicting binary links within knowledge graphs and lack specific tailoring for identifying causal links. The existing approaches often simplify the modeling of causality into a binary triple.  The recent work has aimed to generate event-related causal knowledge graphs from sources like Wikipedia and Wikidata, incorporating causal predicates like hasCause and hasEffect \cite{hassanzadeh2022causalbuilding}. These graphs represent events as nodes and cause-effect relationships as links, with the objective of predicting future events by analyzing the underlying causes and effects of similar past events. Evaluation of causal link prediction tasks often uses established techniques for knowledge graph link prediction. The causal ontology provides a representation platform for both triple-based and more intuitive hyper-relational graph-based causality representation \cite{causalODP}.   The recent work on incorporating causal AI and causal network concepts into knowledge graph link prediction laid the foundation for causal link prediction with causal weights using weighted KG embedding model \cite{jaimini2024causaldisco}.    

\textit{Hyper-relational knowledge graph link prediction}
The appeal to modelling hyper-relational graphs are motivated from conventional triple-based KG embedding models which simplifies the complex property qualifiers.  The convolutional model incorporates complex triples with k qualifiers (key, value) in one fact \cite{guan2019linkNaLP}. However,  all the qualifier pairs are treated equally and does not distinguish between main triple and relation-specific qualifiers. 

The HyperCausalLP approach proposed in this paper innovatively builds upon learned causal networks by transforming them into a CausalKG. It is among the first to use the prior causal structure knowledge encoded in a causal network which in turn is represented in the causal knowledge graph. It distinguishes between the main and the mediated causal link. This transformation allows for the application of KG techniques to discover additional, previously unrecognized causal links, thereby enriching and expanding the causal network beyond what is possible with traditional methods alone. The HyperCausalLP approach predicts new causal links in a KG utilizing the causal weight and four causal relations, i.e., causes, causedBy, causesType, and causedByType.  

\begin{figure*}[h!]
    \centering
    \includegraphics[width=\textwidth, height=4.5cm]{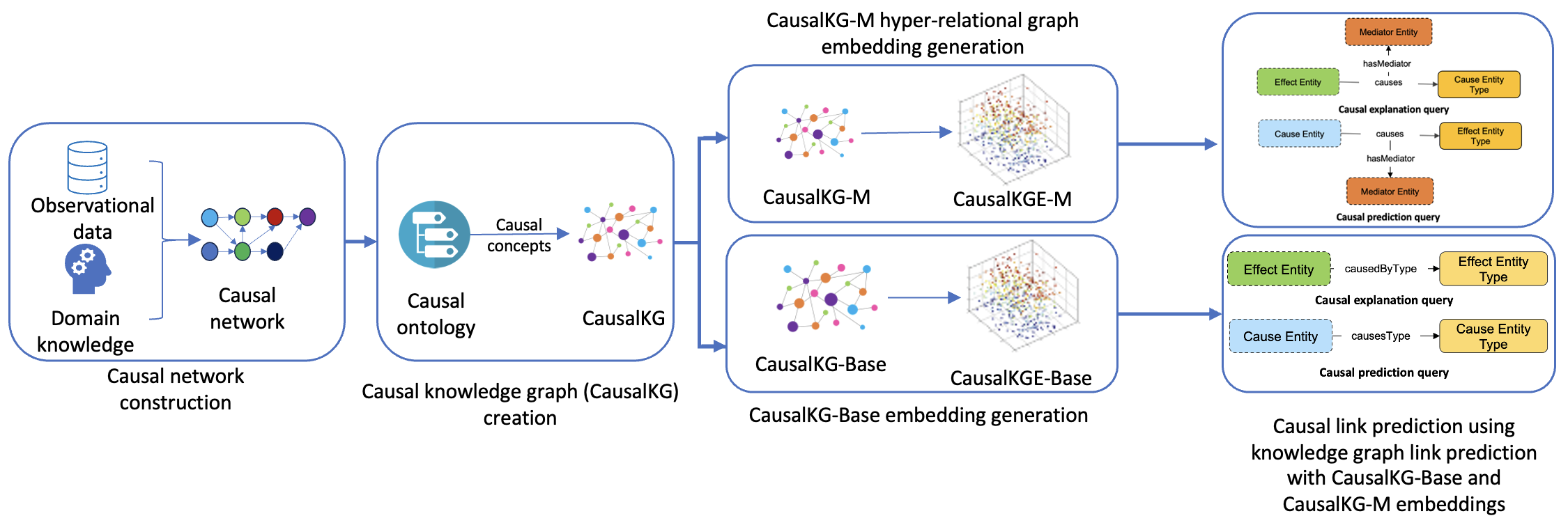}
    \caption{HyperCausalLP has four primary phases: 1) encoding the causal associations in data as a causal network, 2) translating the
causal network into a causal knowledge graph, 3) learning knowledge graph embeddings (CausalKG-Base and hyper-relational graph based embedding CausalKG-M with  mediators as hyper-relations) from the causal knowledge graph,
and 4) using the knowledge graph embeddings for causal link prediction tasks.}
    \label{fig:architecture}
\end{figure*}

\section{Problem Formulation}
The causal link prediction is formulated as a KG link prediction problem. This section defines the primary concepts, including causal relations, causal link, causal entity, qualifier, hyper-relation, and causal knowledge graph. 

\textbf{Causal knowledge graph:}
A causal knowledge graph $CausalKG$ is a hyper-relational KG that includes causal knowledge in the form of causal relations and causal entities. $CausalKG$ = ($N$, $R$, $E$, $E_c$):

 \begin{itemize}
     \item $N$: a set of nodes representing entities
     \item $R$: a set of labels representing relations
     \item $E$ $\subseteq$ $N \times R \times N$: a set of edges representing links between pairs of entities. Each link is a triple $<$$h$, $r$, $t$$>$, where $h$ is the head entity, $r$ is the relation, $t$ is the tail entity.
     \item $N_c \subseteq N$: a set of nodes representing causal entities
     \item $R_c \subseteq R$: a set of labels representing causal relations
     \item $R_m \subseteq R$: a set of labels representing qualifier relations
     \item      $N_m \subseteq N_c$: a set of nodes representing qualifier entities 
     \item $E_c$ $\subseteq$ $N_c \times R_c \times N_c \times P(R_m \times N_m)$: a set of edges representing causal hyper-relation link connecting pairs of causal entities. $P$ denotes the power set.  
\end{itemize}

\textbf{Causal entity: }
A causal entity $n_c \in N_c$ is an entity that is the head or tail of a causal link. There are two types of causal entities: \textit{cause-entity} ($n_{cause}$) and \textit{effect-entity} ($n_{effect}$) such that the \textit{cause-entity} causes the \textit{effect-entity}. However, in the case of a hyper-relation link, causal entity can also be the qualifier entity ($n_m \in N_m$). 

\textbf{Causal relation: }
A causal relation $r_c \in R_c$ is a relation representing a causal association between entities. There are four types of causal relations: 
\begin{itemize}
    \item $causes$ ($r_{causes} \in R_c$) is a causal relation from the cause-entity to the effect-entity.
    \item $causedBy$ ($r_{causedBy} \in R_c$) is a causal relation from the effect-entity to the cause-entity; i.e. the inverse of $causes$.
    \item $causesType$ ($r_{causesType} \in R_c$) is a causal relation from the cause-entity to the type of the effect-entity.
    \item $causedByType$ ($r_{causedByType} \in R_c$) is a causal relation from the effect-entity to the type of the cause-entity.
\end{itemize}

\textbf{Causal link: }
A causal link $e_c \in E_c$ is an edge in the causal KG connecting a pair of causal entities with a causal relation. The causal link is a triple $<$$h_c$, $r_c$, $t_c$$>$, where $h_c$ is the head causal entity, $r_c$ is the causal relation, and $t_c$ is the tail causal entity.

\textbf{Qualifier pair: }
A qualifier pair $q \in Q$ is a hyper-relation in the causal KG connecting a causal link with its hyper-relation relation-entity pair. $Q$ is a set of qualifier pairs{($r_m$, $n_m$)} with qualifier relation $r_m$, and qualifier entity $n_m$. 

\textbf{Qualifier entity: }
A qualifier entity $n_m \in N_m$ is a causal entity that is part of the qualifier pair. 
In a given serial causal connection, the qualifier entities (i.e., mediators) are the entities in between the \textit{cause-entity} and \textit{effect-entity} connected in a sequence, also known as mediators.  In this paper, the qualifier entity refers to the mediator in the serial causal connection. In the context of the paper, the word qualifier entity and mediator can be used interchangeably. 
 
\textbf{Qualifier relation: }
A qualifier relation $r_m \in R_m$ is a relation representing an association between causal link and qualifier entities (or mediator entity). There are two types of qualifier relations: 
\begin{itemize}
    \item $hasMediator$ ($r_{hasMediator} \in R_m$) is a qualifier relation from the causal link to the mediator-entity.
    \item $hasMediatorType$ ($r_{hasMediatorType} \in R_m$) is a qualifier relation from the causal link to the type of the mediator-entity.
\end{itemize}

\textbf{Causal hyper-relational link: }
A causal link $e_c \in E_c$ is an edge in the causal KG connecting a pair of causal entities with a causal relation and their associated mediators. Each causal hyper-relational link is a  tuple $<$$h_c$, $r_c$, $t_c$, $Q$$>$, where $h_c$ is the head causal entity, $r_c$ is the causal relation, $t_c$ is the tail causal entity, $Q$ is a set of qualifier pairs{($r_m$, $n_m$)} with qualifier relation $r_m$, and qualifier entity $n_m$. 

\textbf{Causal link prediction: }
Causal link prediction is the task of finding new causal links in a CausalKG. Given a CausalKG G, this task can be implemented using knowledge graph link prediction. There are two types of distinct causal link prediction tasks: causal prediction and causal explanation.  
\begin{enumerate}
    \item Causal prediction: given a cause-entity ($n_{cause} \in N_c$), the $causesType$ relation ($r_{causesType} \in R_c$), and the qualifier pair ($Q$), find the type ($t$) of the associated effect-entity such that $<$$n_{cause}$, $r_{causesType}$, $t$, $Q$$>$ $\in G$ holds.
    \item Causal explanation: given an effect-entity ($n_{effect} \in N_c$),  the $causedByType$ relation ($r_{causedByType} \in R_c$), and the qualifier pair ($Q$), find the type ($t$) of the associated cause-entity such that $<$$n_{effect}$, $r_{causedByType}$, $t$, $Q$$>$ $\in G$ holds.
  
\end{enumerate}

\section{Methods}
The HyperCausalLP approach is structured into four primary phases (see Figure \ref{fig:architecture}):
(1) finding and encoding the known causal relations into a causal network, (2) translating the causal network into a hyper-relational CausalKG, conformant to the hyper-relational causal ontology incorporating the qualifier pairs (i.e. mediated links), (3) learning hyper-relational KG embedding for the CausalKG, and (4) predicting new causal links in the KG.

\subsection{Causal Network}
A causal network is a graphical model structured as a directed acyclic graph \cite{pearl2009causality}. In this model, nodes represent events, and edges indicate the causal links between these events. The causal network denoted as \(CN = (N^{cn}, E^{cn})\), such that $N^{cn}$ is the set of nodes in the causal network, $E^{cn}$ is the set of edges between nodes.
The direction of each edge in the network indicates the direction of causality.   Given a three-node causal network, the causal links can have three different orientation structures- serial, fork, and collider. A serial structure is one where a causal association is traversed in a series, such as the first event is responsible for causing the second event, and the second event is responsible for causing the third event. In the fork structure,  the first event is responsible for causing both the second and the third event. In the collider structure, two independent events are together responsible for causing the third event.  However, in this paper, we only focus on the serial structure (Figure \ref{fig:causalLink} (A)). The first node is considered a \texttt{cause-entity}, the second node is the \texttt{mediator-entity}, and the third node is the \texttt{effect-entity}.  

\begin{figure}[h!]
\centering
\includegraphics[width=0.6\columnwidth, height=5cm]{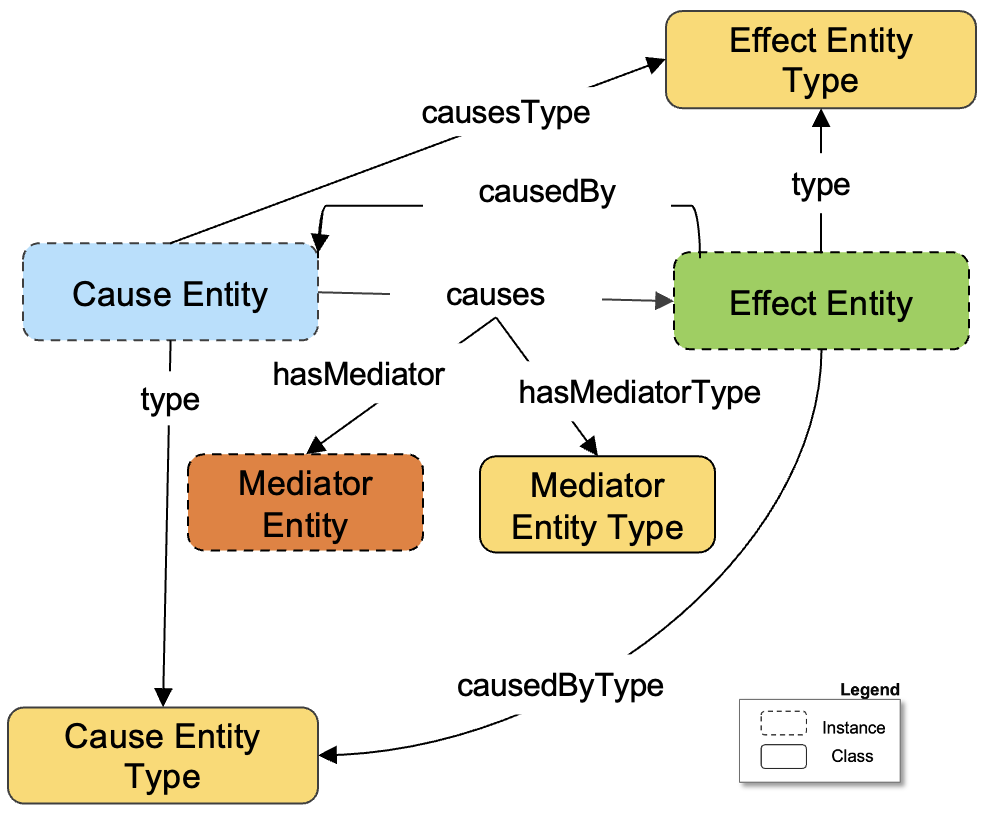}
\caption{The figure shows reified causal relations, causesType and causedByType. The causedByType is a reified relation from an effect-entity instance to the type of a cause-entity. The causesType is a reified relation from a cause-entity instance to the type of an effect-entity. It also illustrates the two qualifier relations associated with causes relation: hasMediator and hasMediatorType. The qualifier relations are also associated with the causedBy relation, which is an inverse of the causes relation.
}
    \label{fig:reifiedRelation}
\end{figure}


\subsection{Hyper-relational Causal Knowledge Graph}
The process of transforming data from a causal network into a hyper-relational causal knowledge graph (CausalKG) involves several straightforward conversions:
\begin{itemize}
    \item $N^{cn} \rightarrow N_c$: nodes in the causal network become causal entities in the CausalKG. The mediator nodes in the causal network become mediator entities in the CausalKG, which are represented as the qualifier entities. 
    \item $E^{cn} \rightarrow E_c$: edges in the causal network become causal links in the CausalKG, of the form 
$<$$n_{cause}$, $r_{causes}$, $n_{effect}$, $r_m$, $n_m$$>$

\end{itemize}


The CausalKG also incorporates other causal relations and qualifier relations such as : $causedBy$, $causesType$,  $causedByType$,  \\ $hasMediator$, and $hasMediatorType$.  The CausalKG consists of all the information from the causal network and is conformant to the hyper-relational causal ontology \cite{causalODP,jaimini2022causalkg}.  The causal ontology is rooted in concepts from causal AI like causal Bayesian networks and do-calculus \cite{causalODP}. It is used to define the semantics and structure of causal relations and the nodes in the causal network.  The ontology defines the primary concepts used to structure a CausalKG, including causal entities, causal relations, and mediators.

The CausalKG is used for causal link prediction using KG link prediction. There are two causal link prediction tasks: causal explanation and causal prediction.  The goal of causal explanation is to predict the type of a cause-entity that is linked to an effect-entity.  The goal of causal prediction is to predict the type of an effect-entity that is linked to a cause-entity. The goal for both tasks is not to predict the specific cause-entity (in the case of causal explanation) or effect-entity (in the case of causal prediction) instance but the type of these respective entities. The cause-entity  (in the case of causal explanation) and effect-entity (in the case of causal prediction) are not directly linked with the cause-entity type and effect-entity, respectively. They are two-hop away: $<$$n_{effect}$, $r_{causedBy}$, $n_{cause}$$>$, $<$$n_{cause}$, $rdf:type$, $type$$>$ for causal explanation; and $<$$n_{cause}$, $r_{causes}$, $n_{effect}$$>$, $<$$n_{effect}$, $rdf:type$, $type$$>$ for causal prediction. The embedding models make predictions about directly linked entities. To overcome the issue of two-hop link prediction, CausalKG uses reified relation (see Figure \ref{fig:reifiedRelation})- 1) for causal prediction: $causeType$ ($r_{causesType} \in R_c$) to add a link connecting a cause-entity with the type of an effect-entity, and 2) for causal explanation: $causedByType$ ($r_{causedByType} \in R_c$) to add a link connecting an effect-entity with the type of a cause-entity. Along with all the above knowledge, the CausalKG also integrates additional domain knowledge associated with the entities that are not distinctly mentioned in the causal network. 

\begin{figure}[h!]
    \centering
    \includegraphics[width=\columnwidth, height=5cm]{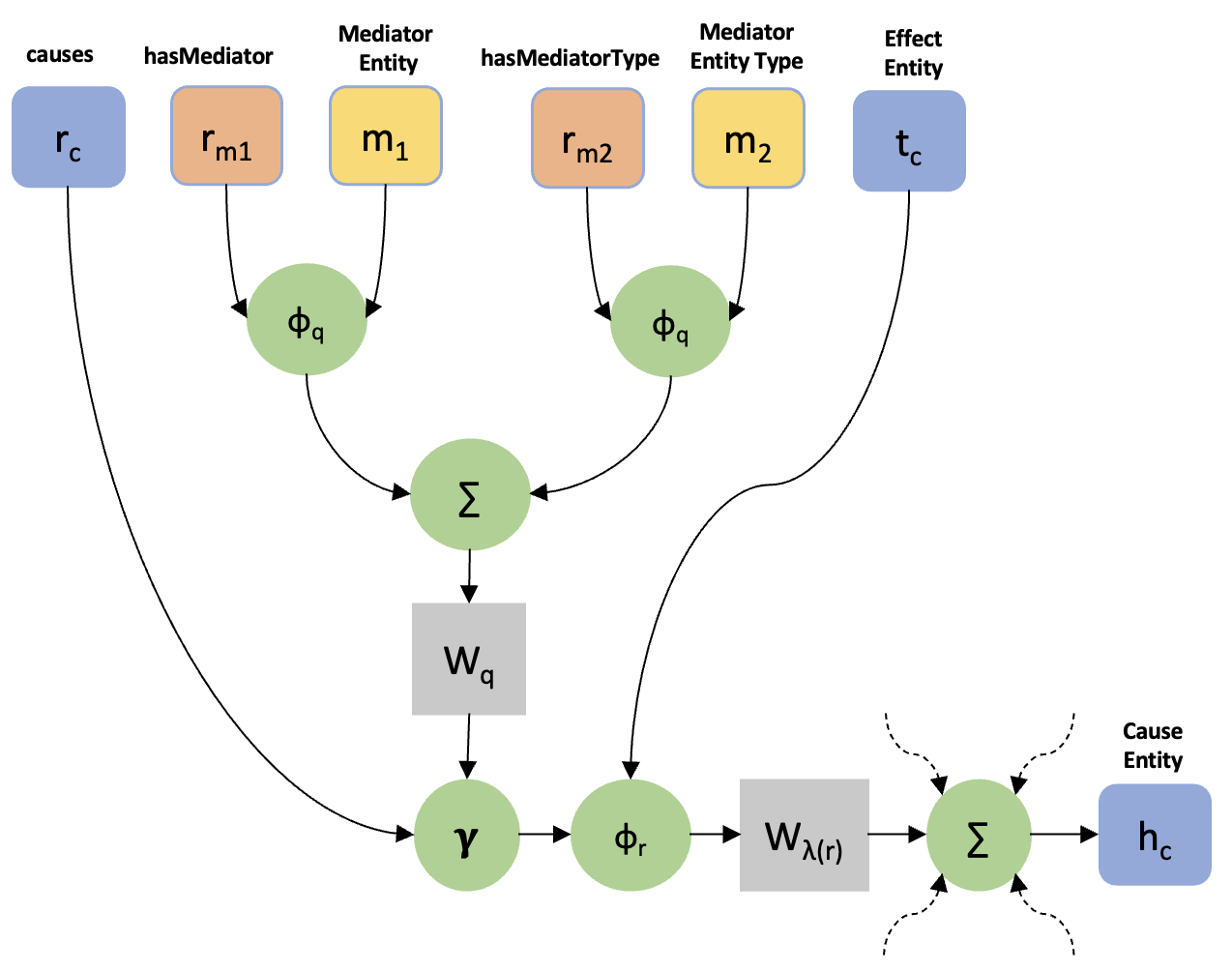}
    \caption{StarE encoder, which encodes a hyper-relations for the causal relation \cite{StarEGalkin2020message}. The hyper-relation qualifier pairs (or mediator pairs) are passed through a composition function $\phi _q$, which are summed together and transformed by weights $W_q$. The transformed vector is merged with $\gamma$ and $\phi _r$. The final node i.e. \texttt{cause entity} combines messages from all the hyper-relation. 
    [Note: As specified in StarE- 1) $\phi$ is a composition function of a node with its respective relation, 2) $W_{\gamma(r)}$ is a direction-specific shared parameter for outgoing, incoming, and self-looping relations, 3) $\gamma$ is a function that combines the main relation, ($r_c$) representation with the representation of
its qualifiers, ($Q$)}
    \label{fig:StarEEncoder}
\end{figure}

\subsection{CausalKG Embedding and Link Prediction}
The CausalKG is converted into a low-dimensional continuous latent vector space representation called KG embeddings (KGE). The KGE is used for downstream tasks such as link prediction, entity classification, triple classification, etc., \cite{wang2017knowledgesurvey}.  The proposed HyperCausalLP approach uses KG embedding algorithms to generate embedding that will be used for causal link prediction. The proposed approach learns two types of KGEs for a CausalKG: 1) CausalKGE-Base embedding without mediators (no hyper-relations), and 2) CausalKGE-M embeddings with mediators as hyper-relations (represented using qualifier pairs). The CausalKGE-Base embedding is trained using the causal links,  ignoring the mediators associated with each link. The CausalKGE-M embedding, on the other hand, is trained using the causal links with the mediators as the hyper-relational links (i.e. qualifiers). The CausalKGE-Base and CausalKGE-M embeddings are evaluated on the task of causal link prediction using KG link prediction. The CausalKG embeddings for  CausalKGE-Base are generated using KG embedding algorithms available in the Ampligraph library \cite{ampligraph}. The CausalKGE-Base uses the four prominent KGE algorithms: TransE \cite{bordes2013translating}, DistMult \cite{yang2014embeddingDistMult}, HolE \cite{nickel2016holographicHolE}, and ComplEx\cite{trouillon2016ComplEx} for embedding generation. The CausalKGE-M is generated a graph neural network based, hyper-relational KGE model, StarE \cite{StarEGalkin2020message}.  

StarE is a graph neural network-based approach.  
It allows a varied number of qualifier pairs to be associated with the causal link. It combines the causal relation embedding with a fixed-length vector representing the associated qualifier pair. It incorporates qualifiers paired with the causal link into the message passing process. The StarE model comprises two parts: a StarE encoder (Figure \ref{fig:StarEEncoder}) and a Transformer decode. The StarE encoder and transformer-based decoder are jointly trained. It initializes two embedding matrices, R (relations) and E (entities). StarE iteratively updates the embedding by message passing across edges in the training set. 
For the task of link prediction, the query is first linearized, and the updated embedding is used to encode the relation and entities. It is then passed through the transformer. The output of the transformer is averaged to get a fixed-dimensional vector representation of the query. The vector is passed through a fully connected layer, multiplied with the entity, and passed through a sigmoid function to obtain probability distribution over all entities. The top n candidate entities for the link prediction query are obtained. 


The proposed approach, HyperCausalLP, formalizes the problem of causal link prediction as a KG link prediction task. The trained CausalKG embedding models, i.e. CausalKGE-Base and CausalKGE-M, are used to predict missing causal links between causal entities in the KG. 
For a given causal link, causal explanation predicts links of form $<n_{effect}, r_{causedByType}, ?, $Q$>$, and causal prediction predicts links of form $<n_{cause}, r_{causesType}, ?, $Q$>$.  For a given dataset with causal entities, causal relations, and mediators associated with the causal links between the entities, HyperCausalLP can be used to create a CausalKG and generate and learn KGE. The generated KGE can be used for causal link prediction. In the next section, we demonstrate and evaluate HyperCausalLP using CLEVRER-Humans, a causal reasoning benchmark dataset \cite{mao2022clevrerhumans}.

\section{Experiments}
The proposed, HyperCausalLP, hyper-relational graph based causal link prediction approach is evaluated using the KG link prediction for two distinct causal link prediction tasks
The above evaluation is demonstrated using a causal benchmark dataset.  
This section details the data, pre-processing steps, creation of a CausalKG from the dataset, experimental setup, evaluation metrics, and description of the evaluation with additional domain knowledge. (Please refer to supplementary for additional details)  

\subsection{Data}
CLEVRER-Humans is a causal benchmark dataset featuring human-annotated causal judgments about physical events depicted in videos \cite{mao2022clevrerhumans}. 
The videos display moving objects that vary in shape (sphere, cube, and cylinder), color (blue, red, yellow, green, purple, gray, cyan, and brown), and material (metal and rubber). Each object can be involved in one of 27 distinct events, such as enter, exit, collide, move, hit, bump, and roll. 
CLEVRER-Humans captures the causal information from these events using a Causal Event Graph (CEG), where the graph's nodes represent event descriptions from the videos and the directed edges indicate causal relationships. The edges of the CEGs are evaluated by human annotators to determine the strength of the causal links between the nodes. These edges are scored on a scale from 1 to 5, where 1 means "not responsible at all," 2 means "a bit responsible," 3 means "moderately responsible," 4 means "quite responsible," and 5 means "extremely responsible.". It is the only large scale causal dataset with 891 causal networks (i.e., CEG)
which provides ground truth for the causal links.

\subsection{Data Pre-processing}
The initial step in generating a CLEVRER-Humans CausalKG involves pre-processing the CEGs.
The CEGs serve as a proxy for a causal network, and their pre-processing is crucial to ensure they align with the definition of a causal network. In a causal network, edges represent causal links between nodes. The first step in this process is to remove edges with a score of 1, indicating no causal responsibility between the two nodes. Next, to maintain the structure of a directed acyclic graph, edges that create cycles in the CEGs are removed. Finally, CEGs are excluded if they do not have any remaining causal links or have a depth of less than 2 from the root node to the leaf node. After pre-processing, we are left with 764 CEGs.

\begin{figure}[h!]
    \centering
    \includegraphics[width=0.8\columnwidth, height=6cm]{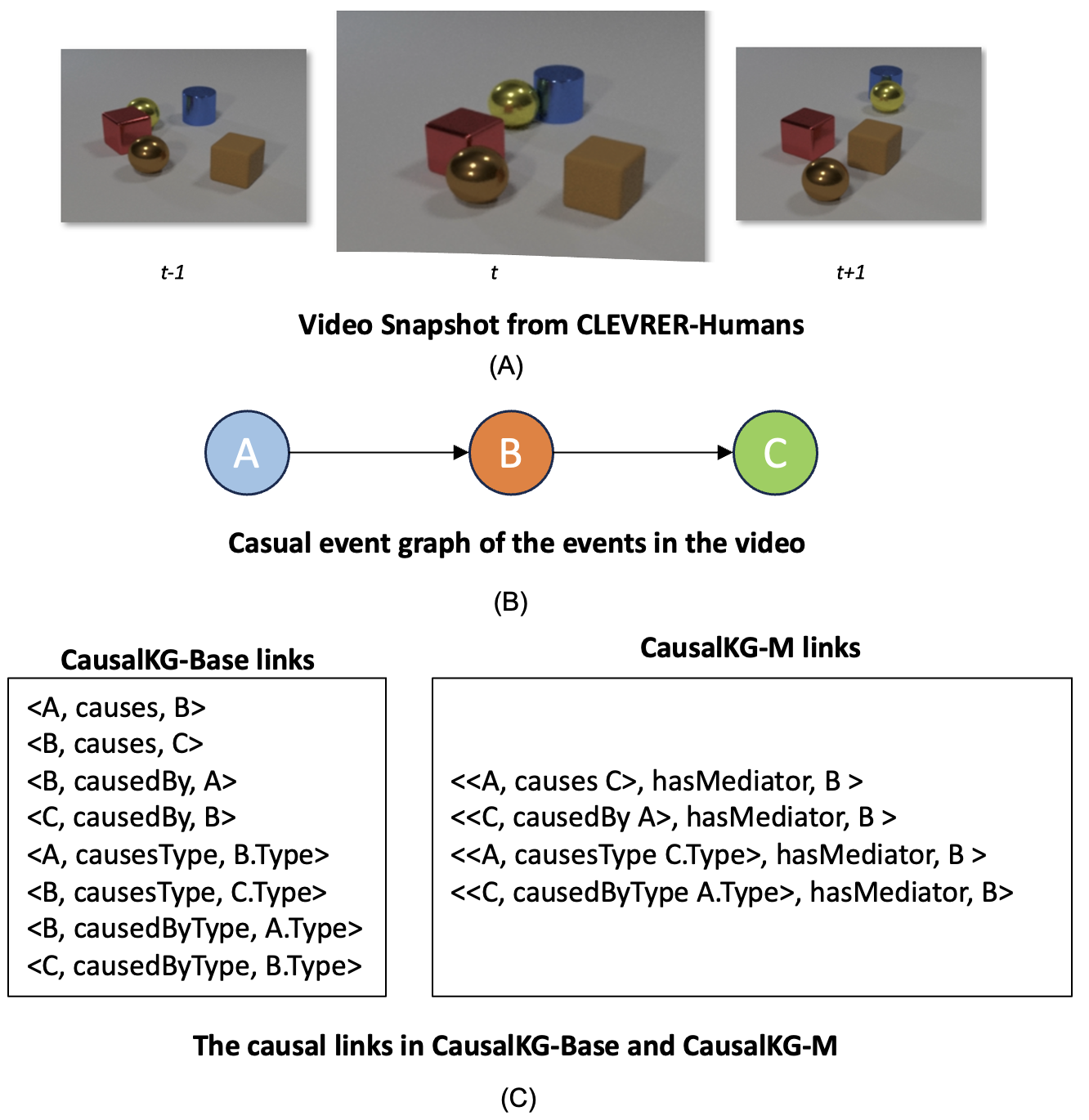}
    \caption{A snapshot of the CausalKG-Base and CausalKG-M representation. (A) A snapshot of collision events in a video at time t-1, t, and t+1 from the CLEVRER-Humans. There are three consecutive collision events that occur: A: the red cube collides with the yellow ball, B: the yellow ball hits the blue cylinder, and C: the blue cylinder moves. The A, B, C are causal entities. A.Type is Collide, B.Type is Hit, and C.Type is Move. (B) The causal event graph of the above snapshot. (C) The causal and mediator (qualifier pairs) links representation in the two different CausalKG.
}
    \label{fig:CausalLinks}
\end{figure}

\subsection{Hyper-relational CausalKG} 
A hyper-relational CausalKG is created from CLEVRER-Humans by encoding the causal information within the CEGs in RDF\footnote{https://www.w3.org/RDF/} format, adhering to the causal ontology. The proposed approach creates two different KG: CausalKG-Base and CausalKG-M (Figure \ref{fig:causalLink}). The CausalKG-Base is a simple KG with causal links, whereas CausalKG-M is a hyper-relational KG, which consists of mediators as hyper-relations (qualifiers). The hyper-relation with the mediator information between two given nodes in the CEG is encoded using RDF-star format. The KG not only includes causal relationships but also details about events (such as hit, collide, push, etc.), the involved objects, and their attributes. CEGs serve as graphical representations of events in the videos. To represent information from the CEGs, we utilize three ontologies: the causal ontology, the scene ontology (prefixed with "so:"), and the semantic sensor network ontology (prefixed with "ssn:"). The causal ontology is employed for events (as causal entities), causal relations, and their corresponding causal mediators (i.e., qualifier pairs). The scene and sensor ontologies depict additional video information, such as scenes, objects, and object characteristics \cite{wickramarachchi2021knowledge, taylor2019semantic}. Each video is depicted as a scene (so:Scene) using scene ontology concepts. This includes representing and connecting the events within the scene (using the so:includes relation), the objects involved (using the so:hasParticipant relation), and the object characteristics (using the ssn:hasProperty relation) \cite{wickramarachchi2021knowledge}. In total, the CausalKG from CLEVRER-Humans contains $>$48K links, 5664 entities, 31 entity types, and 10 relations.

\subsection{Diversifying the Available Knowledge}
\begin{figure}[h!]
 \centering
\includegraphics[width=\columnwidth, height=3.5cm]{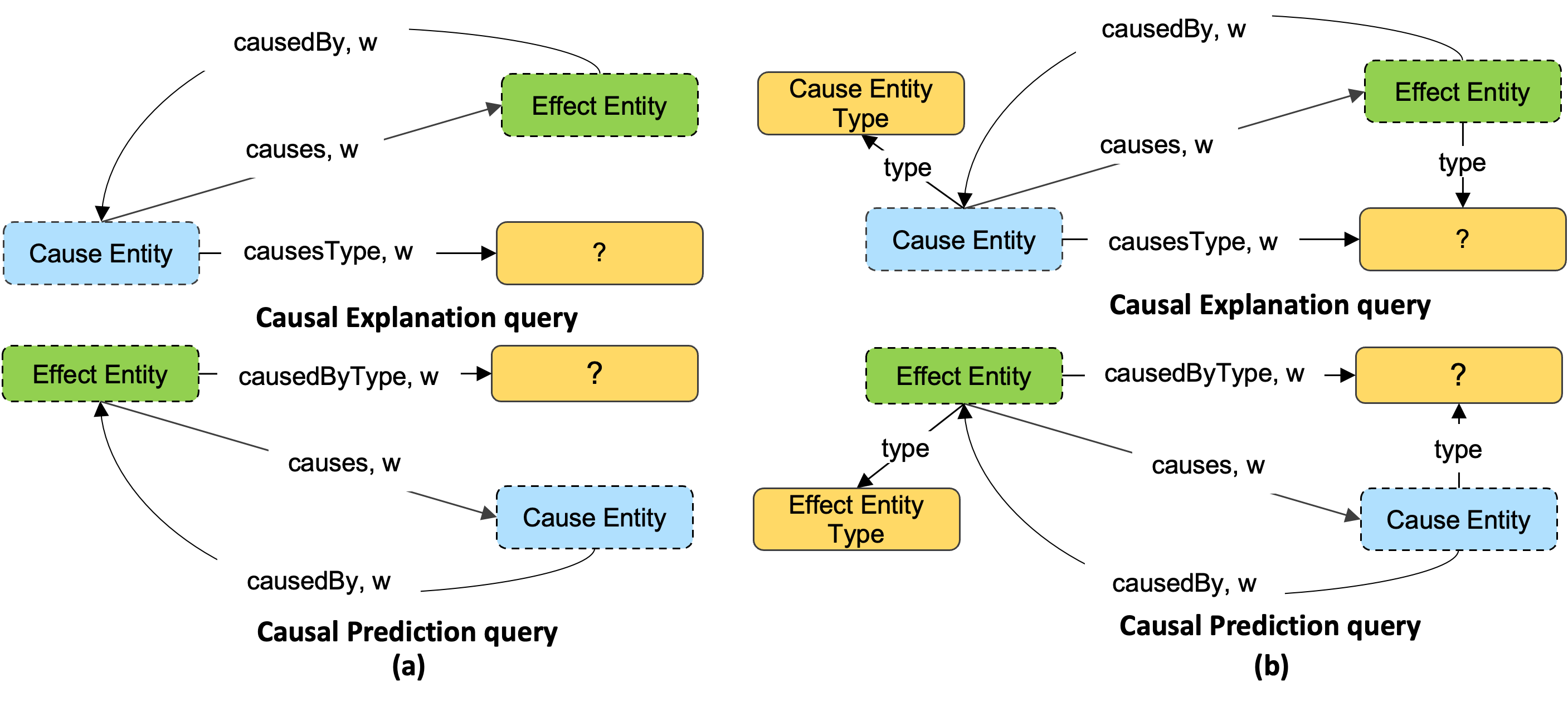}
 \caption{CausalKG structures with additional knowledge (a) subgraph C which consists of links with only causal relations, i.e. \textit{causes}, \textit{causedBy}, \textit{causesType}, and \textit{causedByType}, (b) subgraph CT with causal relations and information about entity types, i.e. \textit{rdf:type}
 In the case of CausalKG-M, the hyper-relations (qualifier pair) are associated with causes, and causedBy causal relation.} 
 \label{fig:KGstructures}
\end{figure}

The CausalKGE-Base and CausalKGE-M embeddings are generated and evaluated on different CLEVRER-Humans CausalKG subgraph structures for the tasks of causal explanation and causal prediction, as illustrated in Figure \ref{fig:KGstructures}. 
In the case of CausalKG-M and the given subgraph, the hyper-relations (qualifier pairs) are associated with causes, and causedBy causal relation as shown in Figure \ref{fig:reifiedRelation}. 
Various graph structures are utilized to assess the performance of HyperCausalLP when different types of information are available in the CausalKG. Specifically, two distinct sub-graph structures are defined with increasing levels of expressivity.
1. The first graph structure, \textbf{C}, shown in Figure \ref{fig:KGstructures}(a), contains only links with causal relations.
2. The second graph structure, \textbf{CT}, shown in Figure \ref{fig:KGstructures}(b), includes links with causal relations and causal entity types (i.e., rdf:type).
We optimized the hyper-parameters for each of these graph structures for causal link prediction tasks i.e., causal explanation and prediction tasks. The CausalKGE-Base models for each graph structures are trained on their respective optimized hyper-parameters (Please refer to supplementary text for more details). The CausalKGE-M model is trained on the StarE hyper-parameters \cite{StarEGalkin2020message}. The trained CausalKGEs are then employed for causal link prediction tasks using well-established link prediction methods.

\subsection{Evaluation Metrics}
HyperCausalLP was evaluated using the KG link prediction for causal link prediction. For a given set of causal links $E_c$ in CausalKG, a set of corrupted links $\mathcal{T}'$ are generated by altering the tail $t_c$ or head $h_c$  of a set of causal links, $<$$h_c, r_c, t_c, Q$$>$, with another causal entity in the KG. Such as replacing the head with $h_c' \neq h_c$ results in $<$$h_c', r_c, t_c, Q$$>$ and replacing the tail with $t_c' \neq t_c$ results in $<$$h_c, r_c, t_c', Q$$>$.  The model assigns scores to the true link $<$$h_c, r_c, t_c, Q$$>$ and corrupted links $<$$h_c', r_c, t_c, Q$$>$, $<$$h_c, r_c, t_c', Q$$>$ $\in \mathcal{T}'$. The scores are sorted to obtain the rank of the true link. The filtered evaluation setting and filtered corrupted links $\mathcal{T}'$ are used to exclude the links present in the training and validation set. The performance of the HyperCausalLP was evaluated using two metrics- Mean reciprocal rank (MRR), and Hits@K (Hits@K, where K=1,3,10). MRR is the mean over the reciprocal of individual ranks of the test links. Hits@k is the ratio of test links present among the top k-ranked links. The higher values of both metrics signify the better performance of the model.  The experiments are performed on a server with NVIDIA TESLA V100 GPU (32 GB GPU memory) and Intel Xeon Platinum 8260 CPU @2.40GHz.

\begin{table}[h!]
\centering
\includegraphics[width=\columnwidth]{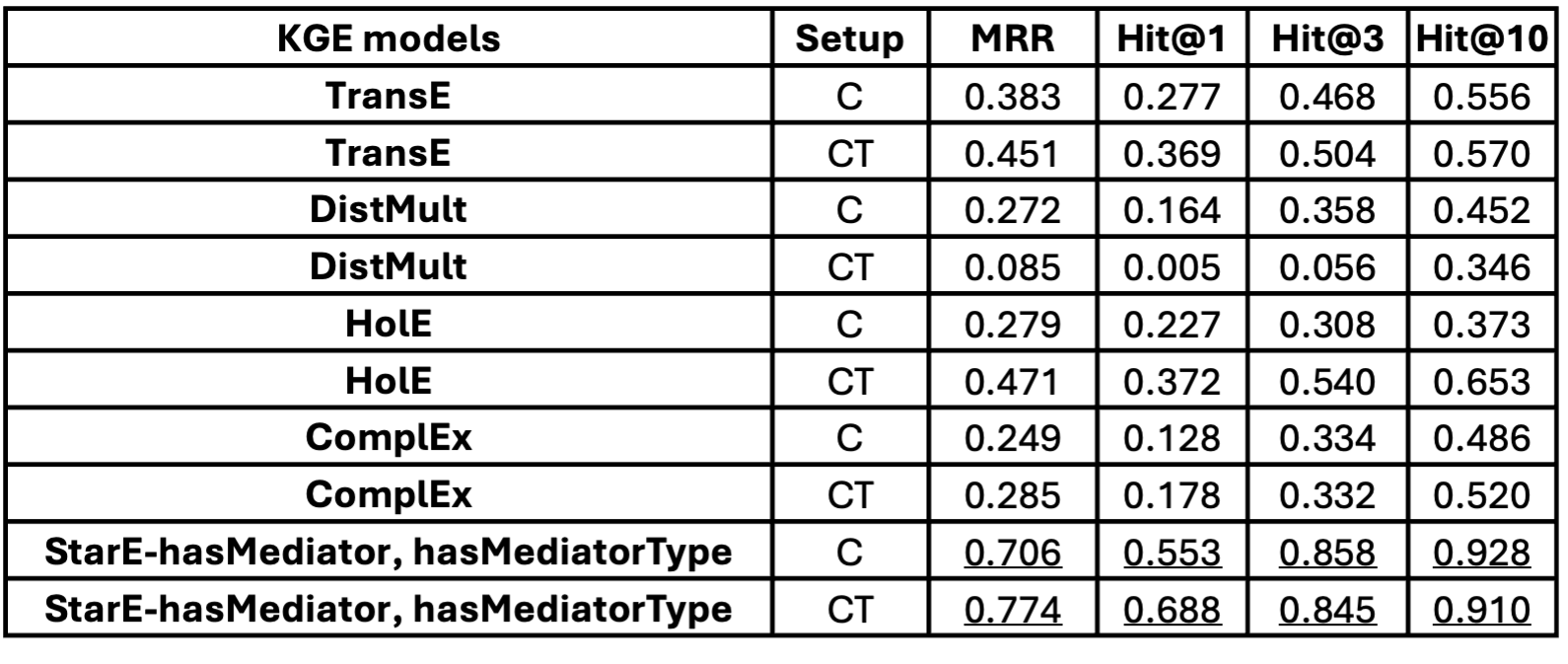} 
\caption{Evaluation metric results for causal prediction for different subgraph structures as shown in Figure \ref{fig:KGstructures}. Hyper-relational CausalKG model with mediator (StaE) show significantly improved performance (underlined) in both mean reciprocal rank (MRR) and Hits@k.}
\label{tab:prediction}
\end{table}

\begin{table}[h!]
\centering
\includegraphics[width=\columnwidth]{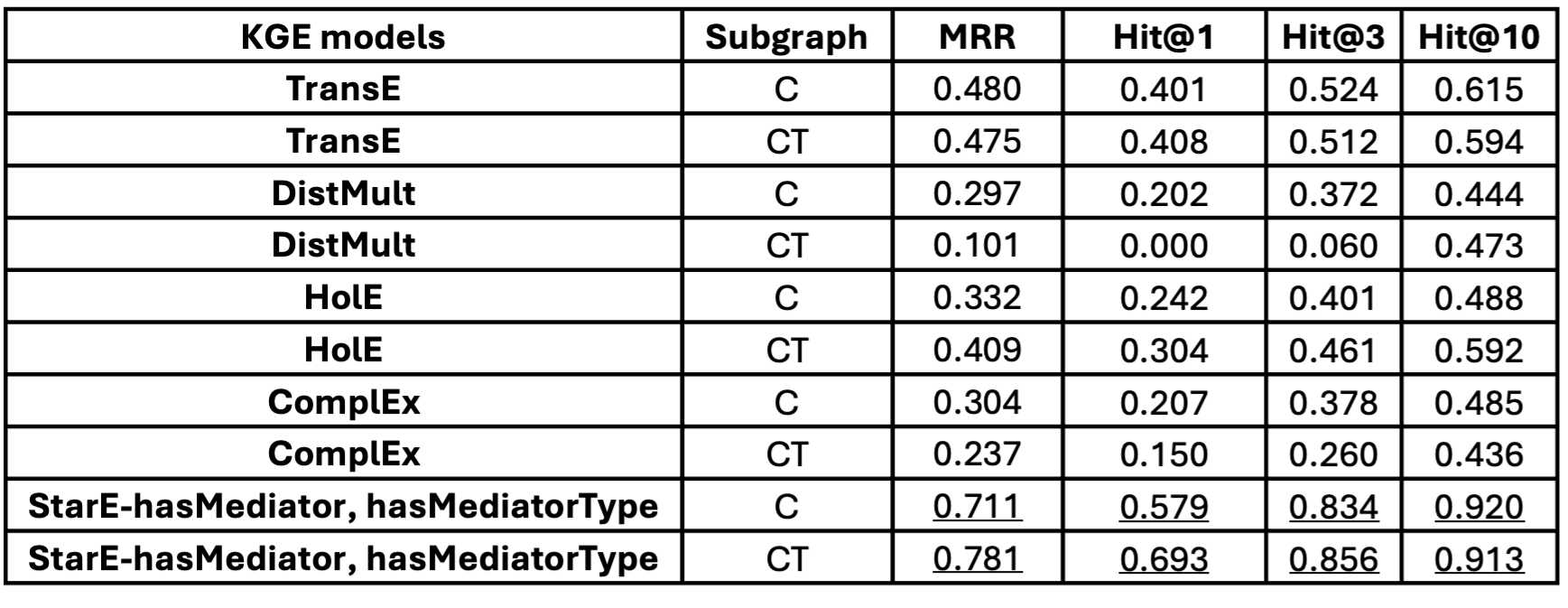} 
\caption{Evaluation metric results for causal explanation for different subgraph structures as shown in Figure \ref{fig:KGstructures}. Hyper-relational CausalKG model with mediator (StaE) show significantly improved performance (underlined) in both mean reciprocal rank (MRR) and Hits@k.}
\label{tab:explanation}
\end{table}

\section{Results and Discussion}
To evaluate HyperCausalLP, we first transformed the CEGs (i.e., causal network) in the CLEVRER-Humans dataset to a hyper-relational CausalKG (\textbf{\textit{RQ1}}). 
The causal links in the hyper-relational  CausalKG preserve the structure of the causal relations. The hyper-relational CausalKG from CLEVRER-Humans is then transformed into KG embeddings. We consider two types of embeddings: baseline embeddings (i.e., without mediators as hyper-relations) and mediated embeddings. HyperCausalLP was evaluated on CausalKG generated from the CLEVRER-Humans dataset for causal link prediction tasks using the trained KG embeddings (RQ2). 
The approach was evaluated on CausalKG-M with StarE with two hyper-relations (i.e., hasMediator and hasMediatorType) along with different CausalKG subgraphs (Figure \ref{fig:KGstructures})

Table \ref{tab:explanation}, Table \ref{tab:prediction} shows the performance of MRR and Hit@K(k=1,3,10) for five KGE models evaluated on different CausalKG subgraph which demonstrate the use of additional knowledge. The results (i.e MRR, Hit\@K) shows a significant increase in the performance of CausalKG-M over CausalKG-Base, the baseline models with no hyper-relations (or mediator information) and just links (\textbf{(RQ3)}). The CausalKG-Base was evaluated with four KGE models- TransE, DistMult, HolE, and ComplEx. The incorporation of additional knowledge (i.e., CT) in the CausalKG-M across different mediator setup shows improved MRR performance over the simpler C subgraph by 5.47\% on average for the causal link prediction tasks (\textbf{(RQ4)}). 
The incorporating mediators with causal link provides an additional knowledge which is crucial for the causal link prediction task. The hyper-relation, hasMediator and hasMediatorType performs the best comparing the MRRs and Hit@k across the board. 
We successfully demonstrated the knowledge incorporated in the hyper-relations (qualifies) significantly improves the causal link prediction.

\section{Conclusion}
The paper introduced an approach to finding missing causal link in an incomplete causal network. The HyperCausalLP, a hyper-relational KG based causal link prediction using KG prediction. The proposed method incorporates the mediator information from the CBN as a hyper-relation in the KG.  The KGE models trained with qualifier (mediator, or hyper-relations) outperform all baseline KGE metrics without qualifiers. The results demonstrate that an effective fusion of causal links with qualifier (mediator, or hyper-relations) in a KG can facilitate the completion of incomplete causal network. Future work will investigate incorporating a varied number and type of mediators as hyper-relations, which will allow multi-hop causal entity prediction. We would also like to extend the HyperCausalLP with a selection of hyper--relational KG embedding models.  

\bibliography{aaai25}
\appendix

\end{document}